\documentclass[letterpaper]{article}
\usepackage[letterpaper]{geometry}

\usepackage[utf8]{inputenc}
\usepackage[T1]{fontenc} 
\usepackage[cjk]{kotex}
\usepackage{amsmath}
\usepackage{graphicx}
\usepackage{url}
\usepackage[table]{xcolor}

\usepackage{latexsym}

\usepackage{hyperref}
\definecolor{darkblue}{rgb}{0, 0, 0.5}
\hypersetup{colorlinks=true,citecolor=darkblue, linkcolor=darkblue, urlcolor=darkblue}

\usepackage{tabularx}
\usepackage{arydshln}
\usepackage{booktabs}

\usepackage{setspace}
\usepackage{lineno}

\usepackage[authoryear,round]{natbib}
\usepackage{algorithmic,algorithm}
\renewcommand{\algorithmiccomment}[1]{\bgroup\hfill$\triangleright$~#1\egroup}

\usepackage[linguistics]{forest} 
\usepackage{synttree}
\usepackage{subcaption}
\usepackage{tikz-dependency}

\usepackage{langsci-gb4e}

\title{Refining Word-Based Grammatical Error Annotation for L2 Korean}

\author{ 
{Jungyeul Park}$^{1\dagger}$\thanks{Corresponding authors: Jungyeul Park and Jayoung Song. Email: \url{jungyeul@kaist.ac.kr} and \url{jayoung.song@psu.edu}. $^{\dagger}$Jungyeul Park, KyunTae Lim, and Wonjun Oh contributed equally to this work.}~~ {Kyungtae Lim}$^{1}$$^{\dagger}$~~ {Wonjun Oh}$^{1}$$^{\dagger}$\\
{Benjamin Nguyen}$^{2}$~~ {Zihao Huang}$^{2}$~~ {Mengyang Qiu}$^{3}$~~{Jayoung Song}$^{4}$$^{*}$\\
$^{1}${Korea Advanced Institute of Science \& Technology, South Korea.}\\
$^{2}${The University of British Columbia, Canada}\\
$^{3}${Saint Elizabeth University, USA}~~
$^{4}${The Pennsylvania State University, USA}
}
\date{ 
}

\begin{document}
\maketitle
\begin{abstract}
Korean grammatical error correction (K-GEC) presents a structural mismatch between word-based evaluation and the morpheme-level locus of many learner errors. Postpositions and verbal endings are bound to lexical hosts, but they encode grammatical relations that must be represented in correction and evaluation. This paper refines word-based grammatical error annotation for L2 Korean by addressing three connected problems in existing resources: surface target realization, Korean-specific edit annotation, and single-reference evaluation. We reconstruct target sentences from the National Institute of Korean Language (NIKL) L2 corpus under morphologically constrained realization rules and convert its morpheme-level annotations into word-level \texttt{m2} edits. We then define a Korean ERRANT-style annotation scheme that preserves the MRU core while distinguishing functional morpheme errors, spelling errors, word boundary errors, and word order errors. We also augment the KoLLA corpus with an additional reference correction, yielding a multi-reference evaluation setting for Korean GEC. Empirical validation shows that the refined NIKL targets yield lower perplexity, the converted \texttt{m2} files achieve higher agreement with source-target edit representations, and the refined resources improve KoBART-based correction under the same model setting. Multi-reference KoLLA evaluation further reduces the penalty imposed on valid corrections that diverge from a single reference, especially for neural and prompted GEC systems. These results show that Korean GEC evaluation depends not only on correction models, but also on reference data and edit annotations that reflect Korean morphology, spacing, and correction variability.
\end{abstract}

\tableofcontents
\doublespacing

\section{Introduction}

Korean grammatical error correction (K-GEC) exposes a structural mismatch between the linguistic form of learner errors and the units assumed by standard GEC evaluation. Many Korean learner errors are located inside the eojeol, where lexical stems combine with bound functional morphemes such as postpositions and verbal endings. These morphemes encode case, clause linkage, tense-aspect, modality, politeness, and discourse-sensitive relations, but word-based GEC treats the correction target as a sequence of surface tokens. The central problem is therefore not only how to correct learner Korean, but how to represent Korean corrections so that word-level evaluation remains linguistically meaningful.

Existing Korean GEC resources inherit this mismatch in two ways. The NIKL L2 learner corpus provides rich morpheme-level correction annotation, but does not directly provide surface target sentences suitable for word-based training and evaluation \citep{yoon-etal-2023-towards}. Because Korean surface realization is not the inverse of morphological segmentation, target reconstruction becomes an empirical and linguistic decision, rather than a mechanical concatenation step. At the same time, previous word-based error labels collapse many Korean-specific distinctions into broad categories such as particle, ending, or spelling errors. Such labels identify the affected domain, but they do not represent the relation between a lexical host and its functional morphology.

A second limitation concerns reference multiplicity. Korean learner sentences often admit more than one grammatical correction because argument omission, case marking, verbal endings, word order, lexical choice, and discourse recoverability interact in the realization of meaning. A single reference therefore cannot exhaust the space of valid corrections. This is especially relevant for neural and prompted GEC systems, which may produce fluent corrections that diverge from the sole reference while preserving grammaticality and intended meaning.

This paper refines word-based grammatical error annotation for L2 Korean by addressing these representational bottlenecks. 
First, we reconstruct surface target sentences from the NIKL L2 corpus under morphologically constrained realization rules, showing that target reconstruction is a substantive part of Korean GEC resource construction rather than a mechanical detokenization step. 
Second, we convert the original morpheme-level annotations into word-level \texttt{m2} edits and define a Korean ERRANT-style scheme that preserves the MRU core of \texttt{ERRANT} while making Korean functional morphology visible in the edit label \citep{bryant-etal-2017-automatic,bryant-2019-automatic}. 
The scheme distinguishes functional morpheme errors, spelling errors, word boundary errors, and word order errors, thereby retaining compatibility with word-level evaluation without treating Korean word forms as morphologically opaque strings.

We further augment the KoLLA corpus with an additional reference correction, creating, to our knowledge, the first multi-reference evaluation setting for Korean GEC \citep{dickinson-israel-lee-2010-building,lee-dickinson-israel-2012-developing}. The augmented resource captures correction variability arising from lexical ambiguity, contextual enrichment, and minimal sentence-level intervention. It also situates Korean GEC within broader multi-reference evaluation practices developed for Chinese and multilingual GEC \citep{zhang-etal-2022-mucgec,masciolini-etal-2025-multigec}.

The empirical results show that these representational refinements affect Korean GEC evaluation at both the resource and model levels. 
The reconstructed NIKL targets yield lower perplexity across Korean language models and substantially higher reconstruction accuracy for contracted and variant eojeols. 
The refined \texttt{m2} conversion better preserves the correction content of the original morpheme-level annotations, improving agreement with source-target edit representations. 
In downstream evaluation, KoBART+ improves over the previous KoBART baseline under the same model setting, indicating that better reconstructed targets provide a cleaner supervision signal. 
Multi-reference KoLLA evaluation further shows that many valid Korean corrections are missed under a single-reference setup, especially for neural and prompted GEC systems. 
These results support the central claim of the paper: Korean GEC evaluation depends not only on correction models, but also on reference data and edit annotations that respect Korean morphology, spacing conventions, and correction variability.

\section{Background and challenges in Korean GEC}
\label{challenges-k-gec}

Korean grammatical error correction is shaped by a representational tension between the surface word, which is the usual unit of neural correction and edit-based evaluation, and the morpheme, which is the locus of many grammatical contrasts in Korean. Case markers, verbal endings, honorific morphology, tense-aspect markers, and clause-linking endings are expressed as bound morphemes, but learner sentences and system outputs are normally evaluated as word-level strings. This mismatch makes Korean GEC strongly dependent on how learner corpora are reconstructed, how morpheme-level corrections are projected onto surface forms, and how errors are typed for evaluation. Existing K-GEC resources therefore raise three connected problems: the status of learner corpora as correction data, the recovery of surface target sentences from morpheme-level annotation, and the compatibility of Korean-specific errors with standard GEC evaluation.

\subsection{Korean learner corpora for GEC}
\label{learner-corpora-kgec}

Previous work on K-GEC has relied on two types of data: L1 writing from the Center for Teaching and Learning for Korean and L2 writing from the National Institute of Korean Language corpus \citep{yoon-etal-2023-towards}. Both datasets include original sentences, corrected sentences, and automatically generated error-annotated files in an ERRANT-style \texttt{m2} format. A standard 70/15/15 split for training, validation, and testing has been proposed.

The two datasets differ substantially in their linguistic status. The L1 dataset consists of sentences drawn from Korean education materials and dictionary examples provided by NIKL. Since these sentences are originally error-free, erroneous sentences were synthesized by transcribing them with Text-to-Speech software and retaining only those in which errors were induced. This design is useful for producing controlled pseudo-error data, but it is less directly informative about the distribution of errors in authentic learner writing.

By contrast, the L2 dataset is based on learner-produced Korean. NIKL collected 3,613 files containing original learner sentences, their morphological segmentation with part-of-speech tags, and morpheme-level error annotations. 
Each correction is represented as an operation on the relevant morphemes, including addition, deletion, and replacement. A representative case is given in Figure~\ref{correction-example}, where the learner form 나왔을 \textit{nawasseul} is analyzed as 나오+았+을 \textit{nao-ass-eul} (`come+\textsc{past}+\textsc{rel}') and corrected as 올라가+면+\textsc{null} \textit{ollaga-myeon}-null (`rise+\textsc{if}’), without an explicitly provided target surface form.

\begin{figure}[!ht]
\centering
{
\footnotesize{
\begin{tabular}{l rl c} \toprule
\multicolumn{2}{r} {Original} & Correction &  \\ 
\textsc{word}& \textsc{morph} & \textsc{morph} &  annotation \\\midrule
 나왔을 \textit{nawasseul} & 나오 \textit{nao} (`come')&  올라가 \textit{ollaga} (`raise') & \textsc{replace}\\
 &  았 \textit{ass} (`\textsc{past}')& 면 \textit{myeon} (`\textsc{if}')& \textsc{replace}\\
&  을 \textit{eul} (`\textsc{rel}')&  *NULL* & \textsc{delete}\\ 
\bottomrule
\end{tabular}
}
}
\caption{Example of morpheme-level correction in the NIKL L2 corpus, where surface-sentence corrections are not directly provided}
\label{correction-example}
\end{figure}

The NIKL L2 corpus is therefore the more relevant resource for learner-oriented K-GEC, since it reflects naturally occurring learner errors rather than automatically induced pseudo-errors. Its annotation, however, was not designed as a directly usable word-level GEC corpus. The corpus records correction at the level of morphologically segmented units, while most GEC systems and evaluation protocols require pairs of surface sentences. The central resource problem is thus not only the availability of learner data, but the conversion of linguistically rich morpheme-level annotation into reliable sentence-level correction data.

\subsection{Morpheme-level correction and surface realization}
\label{morpheme-correction-surface-realization}

The morpheme-level design of the NIKL L2 corpus is linguistically appropriate for Korean, but it creates a non-trivial surface realization problem. Korean morphology is agglutinative, yet surface word forms are not obtained by simple concatenation of morphemes. Stem-final segments, vowel contraction, irregular predicates, and orthographic conventions interact in the realization of inflected forms. As a result, morphological segmentation is not simply the inverse of concatenation \citep{tsarfaty-etal-2012-joint}.

The NIKL L2 corpus does not directly provide surface-sentence corrections. Previous work therefore applied a rule-based system grounded in Korean orthographic guidelines to generate target sentences from corrected morpheme sequences \citep{yoon-etal-2023-towards}. This step is indispensable for word-based K-GEC, since training and evaluation require complete source-target sentence pairs. However, rule-based restoration can introduce new errors when orthographic rules are applied outside their proper morphological domain.

Example~\eqref{ortho-error} illustrates such an error in the earlier restoration output. The automatically restored target sentence incorrectly produces 태나서 \textit{taenaseo}, whereas the correct form is 태어나서 \textit{taeeonaseo}, as in Example~\eqref{ortho-error-corrected}.

\begin{exe}
\ex 
\begin{xlist}
\ex \label{ortho-error}
\gll *8월에 태나서 여름이에요 \\
{\color{white}*}\textit{8wol-e} \textit{taenaseo} \textit{yeoreum-ieyo}\\
\ex \label{ortho-error-corrected}
\glll {\color{white}*}8월에 태어나서 여름이에요.\\
{\color{white}*}\textit{8wol-e} \textit{taeeonaseo} \textit{yeoreum-ieyo}\\
{\color{white}*}8month-at born-during summer.\textsc{cop}\\
\trans `(I was) born in August, so it is summer'
\end{xlist}
\end{exe}

Such cases show that target reconstruction is not a merely technical preprocessing step. It determines what counts as the gold correction, and errors in reconstruction directly affect both training data and evaluation references. For Korean GEC, surface realization must therefore be constrained by morphological category, lexical validity, and the distinction between content stems and functional morphemes. In this work, we focus on the L2 dataset and refine the restoration process because authentic learner writing provides the relevant empirical basis for modeling Korean grammatical error correction.

\subsection{Error annotation and evaluation}
\label{error-annotation-evaluation}

The original NIKL L2 annotation is organized along three dimensions. First, each annotation specifies the part of speech of the affected morpheme, distinguishing nouns, verbs, case markers, endings, and other grammatical categories. Second, it records the transformation applied to the morpheme, such as omission, addition, replacement, or misformation.\footnote{The term \textit{misformation} is used in the NIKL annotation to refer to spelling errors, although this usage is not common in the GEC literature.} Third, it may indicate the linguistic dimension of the error, such as pronunciation, syntax, or discourse.

Prior work consolidated these categories into fourteen error types, including \textsc{insertion}, \textsc{deletion}, and \textsc{ws} for word spacing \citep{yoon-etal-2023-towards}. This mapping made the original morpheme-level annotation more compatible with word-based GEC evaluation. Yet it also introduced a loss of linguistic resolution. Since Korean grammatical errors often arise inside the word, especially through case markers, verbal endings, and other bound functional elements, assigning a single word-level label can obscure the internal structure of the correction.

To enforce consistency, prior work introduced two priority rules. The first gives precedence to transformations, ranked as \textsc{insertion} $>$ \textsc{deletion} $>$ others. The second prioritizes error categories in the order \textsc{ws} $>$ \textsc{wo} $>$ \textsc{spell} $>$ \textsc{shorten} $>$ \textsc{punctuation} $>$ others, where \textsc{wo} denotes word order. These rules produce a deterministic label when multiple analyses are possible, but they also collapse complex learner errors into a single category.\footnote{Notably, the \textsc{punctuation} and \textsc{wo} categories do not appear in the generated \texttt{m2} files, suggesting that these error types were not explicitly annotated or effectively utilized.} The resulting annotation is usable for evaluation, but it does not fully expose the Korean-specific distinction between content-word replacement and functional-morpheme correction.

Evaluation raises a related problem. GEC metrics can be broadly divided into edit-based and fluency-based approaches. Early edit-based frameworks such as HOO \citep{dale-kilgarriff-2011-helping} provided character-level annotations and $F_1$ scores. More widely used metrics include the MaxMatch scorer, or $M^{2}$, and I-measure \citep{dahlmeier-ng-2012-better,felice-briscoe-2015-towards}. $M^{2}$ extracts phrase-level edits through edit lattices and reports precision, recall, and F$_{0.5}$, reflecting the field's emphasis on precision. I-measure aligns the source, system output, and reference, and computes weighted accuracy together with an improvement score that measures whether a system improves upon the original learner sentence.

Fluency-oriented metrics evaluate the corrected sentence more globally. GLEU \citep{napoles-etal-2015-ground} adapts BLEU for grammatical error correction by incorporating the source sentence: it rewards $n$-grams that appear in the reference but not in the source, while penalizing erroneous $n$-grams retained from the source. GLEU$^{+}$ \citep{napoles-etal-2016-gleu} improves robustness in multi-reference settings and more directly penalizes uncorrected errors.

For Korean, prior work used both edit-based and fluency-based evaluation. Yoon et al. (2023) introduced KAGAS, which automatically generates \texttt{m2} files encoding edits between source and reference sentences for Korean learner data. This made it possible to compute M$^{2}$-style precision, recall, and F$_{0.5}$, alongside GLEU. This combination balances local edit overlap with sentence-level fluency, but it remains sensitive to the quality of automatically generated target sentences and to the structure of the derived edit annotation.

The \texttt{ERRANT} framework \citep{bryant-etal-2017-automatic,bryant-2019-automatic} has become the \textit{de facto} standard for interpretable GEC evaluation. Unlike $M^{2}$, which evaluates plain system outputs against gold annotations, \texttt{ERRANT} compares system-generated and gold-generated \texttt{m2} files directly. Edits themselves are therefore the unit of comparison. This makes evaluation more transparent, but also places stronger demands on the linguistic adequacy of the edit classification. For Korean, adopting ERRANT-style evaluation requires an annotation scheme that preserves the MRU distinction while representing Korean-specific phenomena, including functional morpheme errors, word boundary errors, spelling errors, and word order errors.

\paragraph{Distribution of NIKL error types}

The distribution of the original NIKL annotations confirms that Korean GEC is strongly shaped by the interaction between lexical content and bound functional morphology. Table~\ref{error-nikl} reports the most frequent word-level combinations of morpheme errors. In the original annotation, each morpheme-level error is encoded through three fields: grammatical category, error pattern, and an optional error level.\footnote{The original NIKL annotation consists of three levels: grammatical category, pattern, and optional error level. Grammatical categories for content morphemes start with \texttt{C}, and those for functional morphemes with \texttt{F}. Content morpheme categories are inherited from the Sejong corpus, such as \texttt{N*} for nouns and \texttt{V*} for verbs, yielding labels such as \texttt{CNNG} for content common nouns and \texttt{CVV} for content verbs. Functional morpheme categories use abbreviations such as \texttt{AP} for adverbial endings and \texttt{NP} for nominal endings, yielding labels such as \texttt{FAP} and \texttt{FNP}. Error patterns include omission (\texttt{OM}), addition (\texttt{ADD}), replacement (\texttt{REP}), and misformation (\texttt{MIF}). The label \texttt{Removed} is also used for unnecessary material without a full category-pattern-error-level specification. Although the error level is designed to indicate pronunciation, syntax, or discourse, it is rarely annotated.}
The symbol \texttt{0} indicates the absence of an error at the corresponding morpheme or error-level position.

Although the single most frequent annotation involves misformation of common nouns, the dominant recurrent pattern concerns functional morphemes, especially replaced or omitted postpositions and verbal endings. These errors are not ordinary lexical substitutions. A particle or ending is a bound grammatical element attached to a nominal or verbal host, and its misuse changes the grammatical relation expressed by the whole word. The frequency of these patterns shows why Korean GEC evaluation cannot rely only on broad surface labels such as \texttt{PARTICLE} or \texttt{ENDING}. The annotation problem is not simply to name the affected category, but to represent how lexical content and functional morphology jointly determine the corrected word form.

\begin{table}[!ht]
\centering
\footnotesize
\begin{tabular}{cc  c  c}
\toprule
NIKL annotation & Occ. (ratio) & Linguistic interpretation & {\color{gray}Previous} \\
\midrule
CNNG|MIF|0+0 & 2118 (0.0485) & content-word spelling & {\color{gray}\texttt{SPELL}}\\
0+FAP|REP|0 & 1815  (0.0416) & postposition replacement & {\color{gray}\texttt{PARTICLE}}\\
0+FNP|REP|0 & 1793  (0.0411) & postposition replacement & {\color{gray}\texttt{PARTICLE}} \\
0+FOP|REP|0 & 1377 (0.0315) & postposition replacement & {\color{gray}\texttt{PARTICLE}} \\
0+FED|REP|0 & 1107 (0.0253) & verbal-ending replacement & {\color{gray}\texttt{ENDING}} \\
0+FOP|OM|0 & 1082 (0.0248) & postposition omission & {\color{gray}\texttt{PARTICLE}}\\
0+FXP|REP|0 & 1037 (0.0237) & postposition replacement & {\color{gray}\texttt{PARTICLE}} \\
\bottomrule
\end{tabular}
\caption{Most frequent NIKL error annotations, their linguistic interpretation, and their labels in the previous word-based scheme.}
\label{error-nikl}
\end{table}

\section{Refining the NIKL L2 GEC corpus}
\label{refining-nikl-l2}

We refine the NIKL L2 corpus by converting its morpheme-level correction layer into a word-based GEC resource. The refinement consists of three operations: reconstructing surface target sentences, converting original annotations into standardized word-level edits, and validating the resulting \texttt{m2} files against source-target edit representations.

\subsection{Target sentence reconstruction}
\label{target-sentence-reconstruction}

Previous K-GEC datasets adopt a word-based design, using words as the basic unit of correction and annotation \citep{yoon-etal-2023-towards}. 
The original NIKL L2 corpus, by contrast, encodes correction at the morpheme level. 
Each entry contains the learner sentence, its morpheme and part-of-speech analyses, and the corrected morpheme sequence with corresponding part-of-speech tags. 
A surface target sentence must therefore be reconstructed from the corrected morpheme sequence before the data can be used for word-based training or evaluation.

This reconstruction step is not a trivial detokenization procedure. 
Korean surface forms are shaped by the interaction of morpheme boundaries, part-of-speech categories, phonological contraction, spacing, and predicate inflection. 
We therefore reconstruct target sentences by strictly following Korean orthographic norms,\footnote{\url{https://korean.go.kr/kornorms/}} while constraining each realization rule by its morphosyntactic environment. 
In particular, contraction and irregular conjugation rules are applied only when the relevant category, stem shape, and following ending jointly license the surface alternation. 
This category-sensitive restriction is crucial because formally similar morpheme sequences may have different realizations depending on whether the stem is nominal, verbal, adjectival, or lexically specified as belonging to an irregular conjugation class.

Figure~\ref{l2-dataset-sample} illustrates the structure of the original annotation. 
The source contains 온도 나왔을 때 \textit{ondo nao-ass-eul ttae} (`when the temperature came out'), while the corrected morpheme sequence replaces 온도+NULL \textit{ondo+NULL} with 온도+가 \textit{ondo-ga}, replaces 나오+았+을 \textit{nao-ass-eul} with 올라가+면+NULL \textit{ollaga-myeon+NULL}, and deletes 때 \textit{ttae}. 
The well-formed target sentence is not obtained by copying word forms from the annotation, but by realizing the corrected morpheme sequence as 온도가 올라가면 \textit{ondo-ga ollagamyeon} (`when the temperature rises').

\begin{figure*}
\footnotesize
\centering
\resizebox{.98\textwidth}{!}
{
\begin{tabular}{r  ccc ccc ccc c}\toprule
\textsc{source}& 그래서 & 온돌은 & ... &  사용 &  후에 & 온도 & 나왔을 &  때 & 편하다 &. \\
& \textit{geulaeseo} & \textit{ondol-eun} & ... & \textit{sayong} &\textit{hu-e} &  \textit{ondo} & \textit{nao-ass-eul} & \textit{ttae} &\textit{pyeonha-da}  &{.} \\
& so & \textit{ondol}+\textsc{top} & ... & use & after+\textsc{ajt} & temperature & come+\textsc{past}+\textsc{rel} & when & comfortable+\textsc{ind}  & . \\\hdashline
\textsc{source morph}& 그래서& 온돌+은& ... & 사용& 후+에& 온도+NULL& 나오+았+을& 때& 편하+다  & {.}\\
\textsc{target morph}& 그래서& 온돌+은& ... & 사용& 후+에& 온도+가& 올라가+면+NULL& NULL& 편하+다  & {.} \\\hdashline
\cellcolor{gray!20}\textsc{target} &
\cellcolor{gray!20} 그래서 &
\cellcolor{gray!20}온돌은 &
\cellcolor{gray!20}... &
\cellcolor{gray!20}사용 &
\cellcolor{gray!20}후에 &
\cellcolor{gray!20}온도가 &
\cellcolor{gray!20}올라가면 &
\cellcolor{gray!20} &
\cellcolor{gray!20} 편하다  &
\cellcolor{gray!20} .  \\

& \textit{geulaeseo} & \textit{ondol-eun} & ... & \textit{sayong} & \textit{hu-e} & \textit{ondoga} &  \textit{ollaga-myeon} & & \textit{pyeonha-da} & {.} \\
& so & \textit{ondol}+\textsc{top} & ... & use & after+\textsc{ajt} & temperature & rise+\textsc{if} & & comfortable+\textsc{ind} & . \\ 
& \multicolumn{10}{c}{{(`So, \textit{ondol} is comfortable when the temperature rises after use.')}} \\
\bottomrule
\end{tabular}
}
\caption{Example entry from the NIKL L2 K-GEC dataset. Each sentence is annotated at the morpheme and POS levels for both source and target. A well-formed \textsc{target} sentence must be reconstructed from the \textsc{target morph} sequence.}
\label{l2-dataset-sample}
\end{figure*}

Earlier work proposed reconstruction rules based on Korean orthographic norms, especially for irregular verb and adjective conjugations. 
For instance, the sequence 들어오+았 \textit{deuleoo+ass} (`come in+\textsc{past}') is merged into 들어왔 \textit{deuleowass} \citep[p.~6729]{yoon-etal-2023-towards}. 
We retain the basic insight that reconstruction must be morphophonologically informed, but we restrict rule application to contexts in which the relevant morphological conditions are satisfied.

The main source of reconstruction error is the misapplication of realization rules outside their proper morphosyntactic environments. Rules designed for verbal and adjectival inflection may produce illicit forms if they are applied across category boundaries.
For example, 아빠에게 \textit{appa-ege} (`father-\textsc{dat}') can be incorrectly transformed into 아빠게 \textit{appa-ge}. 
This error results from applying Rule 34, which contracts a verbal or adjectival stem ending in the vowel ㅏ \textit{a} or ㅓ \textit{eo} with a following inflectional ending. 
Since 아빠 \textit{appa} (`father') is a noun and 에게 \textit{ege} is a dative postposition, the contraction is morphologically illicit. 
We therefore apply contraction and irregular conjugation rules only to predicates, and only when the following morpheme is an ending that can participate in the relevant alternation.

A further difficulty arises from irregular predicate morphology. 
When morphemes are merged, the same written stem may be compatible with more than one realization pattern, depending on the lexical identity of the predicate. 
This is particularly problematic for homonymous predicates whose lemmas have distinct meanings and distinct inflectional classes. 
For example, 걷다 \textit{geotda} (`clear') follows regular conjugation in the sense of clouds or fog clearing, as in 걷어 \textit{geodeo} (`clear-\textsc{conn}'), but 걷다 \textit{geotda} (`walk') follows \textsc{ㄷ}-irregular conjugation, as in 걸어 \textit{georeo} (`walk-\textsc{conn}'). 
Similarly, 묻다 \textit{mutda} (`bury') is regular, whereas 묻다 \textit{mutda} (`ask') is \textsc{ㄷ}-irregular. 
Other ambiguous cases include regular vs.\ \textsc{ㅂ}-irregular pairs such as 굽다 \textit{gupda} (`bend') vs.\ 굽다 \textit{gupda} (`roast'), and \textsc{르}- vs.\ \textsc{러}-irregular pairs such as 이르다 \textit{ireuda} (`say') vs.\ 이르다 \textit{ireuda} (`reach').

The reconstruction procedure therefore does not apply irregular conjugation rules solely on the basis of the local stem-ending configuration.
Instead, we first identify a limited set of ambiguous predicate lemmas whose realization cannot be determined from the morpheme sequence alone.\footnote{In the NIKL L2 corpus, surface restoration is affected by a small number of lemma-level ambiguities, where the same citation form is associated with distinct lexical entries and distinct inflectional classes. 
The ambiguous lemma 묻다 \textit{mutda} (`bury' or `ask') appears 22 times. 
We restored 20 instances with the ㄷ- \textit{d}-irregular realization and marked 2 instances as unresolved, because the morpheme sequence alone does not always determine the intended lexical entry. It may therefore be difficult to decide whether the surface form should follow regular 묻다 \textit{mutda} (`bury') or ㄷ- \textit{d}-irregular 묻다 \textit{mutda} (`ask'), as in {나에게 묻었다} \textit{naege mudeotda} (`buried it for me') vs.\ {나에게 물었다} \textit{naege mureotda} (`asked me').
The lemma 이르다 \textit{ireuda} (`say' or `reach') appears 3 times; 2 instances were restored with the 러- \textit{reo}-irregular realization, and 1 instance was restored with the 르- \textit{reu}-irregular realization. 
Apart from these predicates, we found no additional lemma-level ambiguities in which lexical identity affected the choice of surface realization.}
For these lemmas, the candidate irregular form is checked against the lexical class of the intended predicate. 
When the annotation or context indicates that the target predicate belongs to the irregular class, the corresponding irregular realization is applied; otherwise, the regular form is preserved. 
Because the number of such ambiguous cases is limited, we manually inspected and corrected the relevant examples in the reconstructed targets. 
This prevents overgeneration of irregular forms while still allowing genuine irregular conjugations to be recovered.

The reconstruction also preserves the alignment information encoded in the original annotation. 
In particular, \texttt{add} and \texttt{null} markers are retained during surface realization, so that insertions and deletions remain traceable when the reconstructed target is converted into word-level edits. 
This preserves the connection between the linguistically rich morpheme-level annotation and the surface sentence pairs required for downstream GEC training and evaluation.

\subsection{Conversion from morpheme-level annotations to word-level edits}
\label{morpheme-to-word-edits}

We generate gold-standard \texttt{m2} files from the human-annotated grammatical errors in the NIKL corpus by mapping each morpheme-level annotation \(\mathcal{A}\) to a standardized word-level edit \(\mathcal{A}'\). The conversion preserves the basic MRU distinction, while representing Korean-specific functional morphology inside replacement labels:
\[
\texttt{(M|U):POS}_{i}
\;\mid\;
\texttt{R:POS}_{i}(\texttt{+POS}_{k})
\rightarrow
\texttt{POS}_{j}(\texttt{+POS}_{k'}).
\]
Here, \texttt{M} and \texttt{U} denote missing and unnecessary material associated with part-of-speech \(\texttt{POS}_{i}\). \texttt{R} denotes replacement, where an original category \(\texttt{POS}_{i}\) is replaced by \(\texttt{POS}_{j}\). When the edit involves bound functional morphology, the label additionally specifies \(\texttt{POS}_{k}\) or \(\texttt{POS}_{k'}\), normalized as \texttt{ADP} for postpositions and \texttt{PART} for verbal endings.

\begin{algorithm}[!ht]
\caption{Conversion of NIKL error annotations into standardized edits}
\label{nikl-algorithm}
\footnotesize
\begin{algorithmic}[1]
\STATE \textbf{Input:} one NIKL error annotation $\mathcal{A}$
\STATE \textbf{Output:} standardized edit $\mathcal{A}'$

\STATE \textbf{function} \textsc{m2\_ErrorAnnotation}($\mathcal{A}$):
\STATE \hspace{1em}\textbf{if} \texttt{F*|ADD} $\in \mathcal{A}$ \textbf{then}
\STATE \hspace{2em}// add functional morpheme $\texttt{POS}_{k'}$
\STATE \hspace{2em}$\mathcal{A}' \gets \texttt{R:POS}_{i} \rightarrow \texttt{POS}_{i}\texttt{+POS}_{k'}$

\STATE \hspace{1em}\textbf{elseif} \texttt{C*|ADD} $\in \mathcal{A}$ \textbf{then}
\STATE \hspace{2em}// added content morpheme
\STATE \hspace{2em}$\mathcal{A}' \gets \texttt{M:POS}_{i}$ 

\STATE \hspace{1em}\textbf{elseif} \texttt{Removed} $\in \mathcal{A}$ \textbf{or} \texttt{*|OM} $\in \mathcal{A}$ \textbf{then}
\STATE \hspace{2em}// deletion, unnecessary or omitted
\STATE \hspace{2em}$\mathcal{A}' \gets \texttt{U:POS}_{i}$ 

\STATE \hspace{1em}\textbf{elseif} \texttt{C*|MIF} $\in \mathcal{A}$ \textbf{then}
\STATE \hspace{2em}// spelling form issues
\STATE \hspace{2em}$\mathcal{A}' \gets \texttt{R:SPELL}$ 

\STATE \hspace{1em}\textbf{elseif} \texttt{F*|MIF} $\in \mathcal{A}$ \textbf{then}
\STATE \hspace{2em}// functional morpheme form issues
\STATE \hspace{2em}$\mathcal{A}' \gets \texttt{R:POS}_{i}\texttt{+POS}_{k} \rightarrow \texttt{POS}_{i}\texttt{+POS}_{k'}$

\STATE \hspace{1em}\textbf{elseif} \texttt{C*|REP} $\in \mathcal{A}$ \textbf{then}
\STATE \hspace{2em}// replacement
\STATE \hspace{2em}$\mathcal{A}' \gets \texttt{R:POS}_{i} \rightarrow \texttt{POS}_{j}$

\STATE \hspace{1em}\textbf{end if}
\STATE \hspace{1em}\textbf{return} $\mathcal{A}'$
\STATE \textbf{end function}
\end{algorithmic}
\end{algorithm}

Algorithm~\ref{nikl-algorithm} implements the conversion. The procedure applies a strict precedence order: insertion and deletion are resolved first, followed by spelling errors and then replacement. This order prevents bound functional material from being incorrectly analyzed as an independent token-level insertion or deletion when it is better understood as a change in the morphological composition of a host word.

The crucial distinction is between content morphemes and functional morphemes. A missing content morpheme, annotated as \texttt{C*|ADD}, can stand as an independent lexical unit and is therefore represented as \(\texttt{M:POS}_{i}\). A missing functional morpheme, annotated as \texttt{F*|ADD}, cannot occur independently. It attaches to a host and changes the morphological composition of the word. We therefore encode it as a replacement:
\[
\texttt{R:POS}_{i}
\rightarrow
\texttt{POS}_{i}\texttt{+POS}_{k'}.
\]
The same logic applies to functional morpheme misformation. A content-morpheme form error, annotated as \texttt{C*|MIF}, is treated as \texttt{R:SPELL}, since the erroneous form can be corrected within the same lexical category. A functional-morpheme form error, annotated as \texttt{F*|MIF}, is treated as replacement of the host-plus-functional configuration:
\[
\texttt{R:POS}_{i}\texttt{+POS}_{k}
\rightarrow
\texttt{POS}_{i}\texttt{+POS}_{k'}.
\]
This representation keeps the edit word-based while preserving the grammatical status of Korean bound morphology.

Figure~\ref{korean-word-m2} contrasts a prior word-based annotation with the refined conversion. In the earlier version, the particle corrections are recorded, but the predicate remains uncorrected, yielding the ungrammatical target 나왔을 때 \textit{nao-ass-eul ttae}. In the refined version, particle errors are represented as replacements involving \texttt{NOUN+ADP}, the predicate is corrected to 올라가면 \textit{ollagamyeon} (`rises'), and the redundant token 때 \textit{ttae} is marked as unnecessary. The resulting target, 온도가 올라가면 \textit{ondo-ga ollagamyeon} (`when the temperature rises'), is both grammatical and semantically appropriate.

\begin{figure*}
\begin{subfigure}[b]{\textwidth}    
\centering
\scriptsize
\begin{tabular}{l}
\texttt{S 그래서 온돌은 사용하기 전에 힘들고 편리하지 않지만 사용 후에 온도 나왔을 때 편하다 .}\\
\texttt{A 3 4|||PARTICLE|||전에는|||REQUIRED|||-NONE-|||0}\\
\texttt{A 9 10|||PARTICLE|||온도가|||REQUIRED|||-NONE-|||0}\\
~\\
\multicolumn{1}{c}{\textsc{target}: *그래서 온돌은 사용하기 전에는 힘들고 편리하지 않지만 사용 후에 온도가 {\color{red!80}나왔을 때} 편하다 .}
\end{tabular}

\caption{Prior word-based annotation of the example sentence}
\label{previous-word-based-m2}
\end{subfigure}

\hfill

\begin{subfigure}[b]{\textwidth}    
\centering
\scriptsize
\begin{tabular}{l}
\texttt{S 그래서 온돌은 사용하기 전에 힘들고 편리하지 않지만 사용 후에 온도 나왔을 때 편하다 .}\\
\texttt{A 3 4|||R:NOUN+ADP -> NOUN+ADP|||전에는|||REQUIRED|||-NONE-|||0}\\
\texttt{A 9 10|||R:NOUN -> NOUN+ADP|||온도가|||REQUIRED|||-NONE-|||0}\\
\texttt{A 10 11|||R:VERB -> VERB|||올라가면|||REQUIRED|||-NONE-|||0}\\
\texttt{A 11 12|||U||||||REQUIRED|||-NONE-|||0}\\
~\\
\multicolumn{1}{c}{\textsc{target}: 그래서 온돌은 사용하기 전에는 힘들고 편리하지 않지만 사용 후에 온도가 {\color{green!80}올라가면} 편하다 .} \\
\multicolumn{1}{c}{(`So, the ondol is difficult and not convenient before use, but it is comfortable when the temperature rises after use.')} 
\end{tabular}
\caption{Refined word-based annotation of the example sentence}
\label{refined-word-based-m2-main}
\end{subfigure}
\caption{Comparison of prior and refined word-based \texttt{m2} annotations}
\label{korean-word-m2}
\end{figure*}

\subsection{Validation of reconstructed targets}
\label{validation-reconstructed-targets}

We validate the reconstructed NIKL targets through two tests: target-sentence fluency and intrinsic surface reconstruction. 
The fluency test compares the targets produced by the previous restoration pipeline with those produced by our reconstruction procedure. 
Perplexity is computed over the complete target set using three Korean language models, \texttt{kanana2-30b-a3b},\footnote{\url{https://huggingface.co/kakaocorp/kanana-2-30b-a3b-base-2601}} \texttt{kormo-10b},\footnote{\url{https://huggingface.co/KORMo-Team/KORMo-10B-base}} and \texttt{solar-10-7b-v1},\footnote{\url{https://huggingface.co/upstage/SOLAR-10.7B-v1.0}} and one multilingual model, \texttt{Qwen3-4B}.\footnote{\url{https://huggingface.co/Qwen/Qwen3-4B}} 
We report corpus perplexity as
\begin{align}
  \mathrm{PPL}_{\mathrm{corpus}}
  =
  \mathrm{exp}\Big(
  -\frac{1}{M}
  \sum_{i=1}^{N}\sum_{t=1}^{T_i}
  \log p_{\theta}(y_{i,t}\mid y_{i,<t})
  \Big),
  \qquad
  M=\sum_{i=1}^{N}T_i ,
\end{align}
where \(T_i\) is the number of target tokens in sentence \(i\) after tokenization.

\begin{table}[!ht]
    \centering
    \footnotesize
    \begin{tabular}{llcc} 
        \toprule
        & & \citet{yoon-etal-2023-towards} & Proposed target \\ 
        \midrule
        Korean model 
        & kanana2-30b-a3b (base) & 163.28 & \textbf{153.99} \\
        & KORMo-10b (base) & 194.70 & \textbf{181.56} \\
        & SOLAR-10.7b-v1 (base) & 25.21 & \textbf{24.97} \\ 
        \midrule
        Multilingual model 
        & Qwen3-4b (base) & 32.68 & \textbf{31.08} \\
        \bottomrule
    \end{tabular}
    \caption{Average perplexity of target sentences from the previous restoration pipeline and from the proposed reconstruction procedure. Lower values indicate greater fluency.}
    \label{target-reconstruction}
\end{table}

Table~\ref{target-reconstruction} shows that the proposed targets have lower perplexity across all four models. 
This indicates that the refined reconstruction procedure does not merely preserve the corrected morpheme sequence, but yields more plausible Korean surface sentences.

We then test surface reconstruction intrinsically on source-side sentences in the original NIKL L2 corpus. 
Because the corpus provides both the surface learner sentence and its morpheme-level analysis, this setting directly evaluates whether a morpheme sequence can be converted back into its attested surface form.

\begin{table}[!ht]
\centering
\footnotesize
\caption{Comparison of surface reconstruction accuracy for nouns, verbs/adjectives, and all eojeols.}
\label{tab:comprehensive_model_comparison}
\begin{tabular}{lccc}
\toprule
& \textbf{Noun} & \textbf{Verb/Adjective} & \textbf{All eojeols} \\ 
\midrule
\citet{yoon-etal-2023-towards} & 97.03\% & 91.34\% & 95.21\% \\
Proposed reconstruction & 99.20\% & 97.85\% & 98.77\% \\
\bottomrule
\end{tabular}
\end{table}

Table~\ref{tab:comprehensive_model_comparison} shows that the proposed procedure improves reconstruction accuracy for nouns, verbs/adjectives, and eojeols overall. 
The gain is most important for predicates, where surface realization is affected by contraction, phonological alternation, irregular conjugation, and category-sensitive rule application. 
This confirms that the improvement is concentrated in the cases where Korean surface forms cannot be obtained by simple morpheme concatenation.\footnote{For contracted or variant forms, the proposed procedure achieves 93.39\% accuracy for nouns, with 466 correct reconstructions out of 499; 98.50\% accuracy for verbs/adjectives, with 6,756 correct reconstructions out of 6,859; and 97.97\% accuracy at the eojeol level, with 7,223 correct reconstructions out of 7,373. 
Because this evaluation is conducted on learner sentences, the input may contain grammatical errors. 
We therefore conduct the same reconstruction test on the KLUE dataset \citep{park-etal-2021-klue}, where the proposed procedure obtains 98.87\% overall accuracy and 94.16\% accuracy on contracted or variant eojeols. 
The previous reconstruction procedure obtains 95.04\% overall accuracy and 61.07\% accuracy on contracted or variant eojeols.}

These results show that the reconstructed targets are both more fluent and more faithful to Korean surface realization. 
This is crucial for K-GEC, since reconstructed targets must function as gold correction strings while remaining anchored in the original morpheme-level annotation.

\section{A Korean ERRANT-style automatic annotation scheme}
\label{korean-errant-style-annotation}

The conversion procedure in Section~\ref{morpheme-to-word-edits} is tied to the original NIKL annotation format. 
For broader evaluation, the same principles must apply directly to arbitrary source-correction pairs, including human references and system outputs. 
We therefore define a Korean ERRANT-style automatic annotation scheme that preserves the MRU core of \texttt{ERRANT} \citep{bryant-etal-2017-automatic,bryant-2019-automatic}, while making explicit the Korean-specific phenomena that are not adequately captured by undifferentiated replacement labels.

The purpose of this automatic annotation is twofold. 
First, it provides a more detailed grammatical description of correction edits, allowing system outputs to be analyzed not only as correct or incorrect, but also in terms of the linguistic phenomena they modify. 
Second, the resulting correction edits can be used directly for edit-based GEC evaluation, where system edits are compared with gold edits to compute precision, recall, and $F$ scores. 
Thus, the annotation scheme serves both as an interpretive layer for grammatical error analysis and as an operational edit representation for evaluation.

The scheme is designed for word-based GEC evaluation, but it does not treat the word as morphologically opaque. 
Korean learner errors often involve bound functional morphemes inside an eojeol, and these morphemes determine case, clause type, tense-aspect interpretation, modality, and connective relations. 
A Korean edit annotation scheme must therefore distinguish surface token replacement from changes in the internal grammatical composition of a word. 
Our annotation scheme follows this principle by combining MRU labels with Korean-specific replacement subtypes.

\subsection{MRU core labels}
\label{mru-core-labels}

The annotation scheme adopts the three-way MRU distinction: \texttt{M}issing, \texttt{R}eplacement, and \texttt{U}nnecessary. \texttt{M} marks material that is absent from the source but required in the correction. \texttt{U} marks material present in the source but removed in the correction. \texttt{R} marks material that remains aligned to a source span but changes in form, category, order, or segmentation.

The general form of the annotation is as follows:
\[
\texttt{(M|U):POS}_{i}
\;\mid\;
\texttt{R:POS}_{i}(\texttt{+POS}_{k})
\rightarrow
\texttt{POS}_{j}(\texttt{+POS}_{k'}).
\]
Here, \(\texttt{POS}_{i}\) and \(\texttt{POS}_{j}\) identify the major lexical or grammatical category involved in the edit. When the edit involves bound functional morphology, \(\texttt{POS}_{k}\) and \(\texttt{POS}_{k'}\) specify the functional component attached to the host word. We use \texttt{ADP} for postpositions and \texttt{PART} for verbal endings.

The scheme imposes a deterministic precedence order. Insertions and deletions are resolved before replacements, because missing or superfluous material must first be separated from changes to existing material. Spelling errors are then identified before broader replacement categories. Functional morpheme errors are treated as replacements over host-plus-functional configurations, since postpositions and endings cannot be evaluated as independent surface words. Word boundary and word order errors are classified only when the relevant string or set-based conditions are met.

\subsection{Functional morpheme errors}
\label{functional-morpheme-errors}

Functional morpheme errors constitute one of the central categories in Korean learner writing. Postpositions attach to nominal hosts and encode relations such as subject, object, topic, dative, locative, and genitive. Verbal endings attach to predicate stems and encode tense, aspect, mood, clause linkage, relativization, sentence type, politeness, and related grammatical distinctions. Because these elements are bound, they cannot be treated as ordinary inserted or deleted words in word-level GEC annotation. As shown in Table~\ref{error-nikl}, the most frequent recurrent NIKL patterns involve replaced or omitted functional morphemes. Functional morpheme errors should therefore be analyzed not as broad particle or ending errors, but as changes in the morphological composition of a host word.

We annotate functional morpheme errors as replacements. If a functional morpheme is missing from a host word, the edit has the form:
\[
\texttt{R:POS}_{i}
\rightarrow
\texttt{POS}_{i}\texttt{+POS}_{k'}.
\]
If an incorrect functional morpheme is replaced by a correct one, the edit has the form:
\[
\texttt{R:POS}_{i}\texttt{+POS}_{k}
\rightarrow
\texttt{POS}_{i}\texttt{+POS}_{k'}.
\]
This notation keeps the edit compatible with word-level \texttt{m2} evaluation while preserving the internal grammatical structure of the Korean word.

The condition for functional morpheme replacement is that the host content word remains constant while the functional component differs. For nominal hosts, the functional component is normalized as \texttt{ADP}; for verbal or adjectival hosts, it is normalized as \texttt{PART}. The following example illustrates a nominal functional morpheme correction, where the noun is retained but the case marking changes.

\begin{center}
{
\footnotesize
\begin{tabular}{l}
\texttt{S 비행기 음식이 안 맞았습니다 .}\\
\texttt{A 1 2|||R:NOUN+ADP -> NOUN+ADP|||음식을|||REQUIRED|||-NONE-|||0}
\end{tabular}
}
\end{center}

An omitted postposition is represented in the same replacement family, because the correction changes a bare nominal host into a host-plus-postposition form.

\begin{center}
{
\footnotesize
\begin{tabular}{l}
\texttt{S <SOURCE SENTENCE WITH BARE NOUN>}\\
\texttt{A <START> <END>|||R:NOUN -> NOUN+ADP|||<CORRECTED NOUN+POSTPOSITION>|||REQUIRED|||-NONE-|||0}
\end{tabular}
}
\end{center}

A verbal-ending error is treated analogously. The predicate stem is preserved, but the functional ending changes.

\begin{center}
{
\footnotesize
\begin{tabular}{l}
\texttt{S <SOURCE SENTENCE WITH INCORRECT VERBAL ENDING>}\\
\texttt{A <START> <END>|||R:VERB+PART -> VERB+PART|||<CORRECTED VERB FORM>|||REQUIRED|||-NONE-|||0}
\end{tabular}
}
\end{center}

\subsection{Spelling, word boundary, and word order errors}
\label{spelling-word-boundary-word-order-errors}

Spelling errors are annotated as \texttt{R:SPELL}. This label is used for content-word form deviations where the correction preserves the lexical or grammatical category but changes the written form. Unlike recent work on Chinese GEC, where spelling errors may be separated into phonetic and shape-based subtypes \citep{gu-etal-2025-improving}, we do not introduce such subcategories for Korean in the present scheme. Korean orthography is alphabetic and featural, and many learner spelling errors involve interactions among phonological representation, morphophonological alternation, and orthographic convention. In practice, phonetic and visual similarity are not always separable in a stable way. We therefore collapse content-word spelling deviations into \texttt{R:SPELL}. Functional morpheme misforms are excluded from this category and treated as functional morpheme replacements, since they alter the grammatical attachment of the host word.

\begin{center}
{
\footnotesize
\begin{tabular}{l}
\texttt{S <SOURCE SENTENCE WITH CONTENT-WORD SPELLING ERROR>}\\
\texttt{A <START> <END>|||R:SPELL|||<CORRECTED WORD>|||REQUIRED|||-NONE-|||0}
\end{tabular}
}
\end{center}

Word boundary errors are annotated as \texttt{R:WB}. They are detected when the source and target differ in tokenization but become identical after removing spaces:
\[
\textsc{Join}(\mathcal{S}) = \textsc{Join}(\mathcal{T}).
\]
This condition captures both word merging and word splitting: a source token may correspond to multiple target tokens, and several source tokens may correspond to a single target token. 
In the current implementation, an ERRANT-based Korean aligner first identifies source--target spans, and word-boundary labeling is then applied to those spans \citep{qiu-etal-2025-multilingual,song-lim-park-2026-enriching}. 
The resulting edit is labeled \texttt{R:WB}, because the relevant correction concerns word-boundary placement rather than lexical substitution.

\begin{center}
{
\footnotesize
\begin{tabular}{l}
\texttt{S <SOURCE SENTENCE WITH WORD BOUNDARY ERROR>}\\
\texttt{A <START> <END>|||R:WB|||<CORRECTED SPACING>|||REQUIRED|||-NONE-|||0}
\end{tabular}
}
\end{center}

Word order errors are annotated as \texttt{R:WO}. They are detected when the source and correction contain the same lexical but differ in linear order:
\[
\textsc{Set}(\mathcal{S}) = \textsc{Set}(\mathcal{T}).
\]

This condition abstracts away from order while preserving lexical identity. It is particularly relevant for Korean because relatively flexible constituent order does not imply unrestricted ordering. Word order interacts with information structure, scope, argument prominence, modifier attachment, and discourse naturalness. A sequence may therefore contain the same words as its correction while still requiring reordering.

\begin{center}
{
\footnotesize
\begin{tabular}{l}
\texttt{S <SOURCE SENTENCE WITH WORD ORDER ERROR>}\\
\texttt{A <START> <END>|||R:WO|||<REORDERED CORRECTION>|||REQUIRED|||-NONE-|||0}
\end{tabular}
}
\end{center}

Figure~\ref{error-example} summarizes the three major replacement subtypes. The examples are schematic and will be replaced with Korean corpus examples in the final version.

\begin{figure}[!ht]
    \centering
{    
\footnotesize{
\begin{tabular}{rl} \hline 
 \textsc{r:spell} & \texttt{<misspelled content word>} $\rightarrow$ \texttt{<corrected content word>} \\
 \textsc{r:wo} & \texttt{<word$_1$ word$_2$ word$_3$>} $\rightarrow$ \texttt{<word$_2$ word$_1$ word$_3$>} \\
 \textsc{r:wb} & \texttt{<word boundary error>} $\rightarrow$ \texttt{<correct spacing>} \\ \hline
\end{tabular}}
}
\caption{Schematic examples of generalized \texttt{R}eplacement error types.}
\label{error-example}
\end{figure}

\subsection{Generalized classification algorithm}
\label{generalized-classification-algorithm}

Algorithm~\ref{error-classification-algorithm-detail} formalizes the annotation procedure. Given a source span \(\mathcal{S}\) and a correction span \(\mathcal{T}\), the classifier returns an edit label from the MRU inventory and the Korean-specific replacement categories. Functional morpheme differences are checked first because they are grammatically central and can be obscured if the word is treated as an opaque string. Spelling mismatches in content words are then identified. Word order and word boundary errors are detected by set equality and space-insensitive string equality, respectively. If none of these Korean-specific conditions is met, the edit is returned under the general MRU classification.

\begin{algorithm}[!ht]
\caption{Generalized error classification}
\label{error-classification-algorithm-detail}
{\footnotesize
\begin{algorithmic}[1]
\STATE \textbf{Input:} source span $\mathcal{S}$ and correction span $\mathcal{T}$
\STATE \textbf{Output:} edit label $\mathcal{A}$
\STATE \textbf{function} \textsc{ErrorClassification}($\mathcal{S}$, $\mathcal{T}$):
\IF{$\mathcal{S} = \emptyset$}
    \RETURN{\texttt{M:POS}$_{j}$}
\ELSIF{$\mathcal{T} = \emptyset$}
    \RETURN{\texttt{U:POS}$_{i}$}
\ELSIF{host content word is identical and functional morpheme differs between $\mathcal{S}$ and $\mathcal{T}$}
    \RETURN{\texttt{R:POS$_{i}$(+POS$_{k}$)} $\rightarrow$ \texttt{POS$_{i}$(+POS$_{k'}$)}}
\ELSIF{spelling mismatch occurs in a content word}
    \RETURN{\texttt{R:SPELL}}
\ELSIF{\textsc{Set}($\mathcal{S}$) == \textsc{Set}($\mathcal{T}$)}
    \RETURN{\texttt{R:WO}}
\ELSIF{\textsc{Join}($\mathcal{S}$) == \textsc{Join}($\mathcal{T}$)}
    \RETURN{\texttt{R:WB}}
\ENDIF
\RETURN{\texttt{R:POS}$_{i}$ $\rightarrow$ \texttt{POS}$_{j}$}
\STATE \textbf{end function}
\end{algorithmic}
}
\end{algorithm}

The resulting annotation is deterministic and corpus-independent. It can be applied to NIKL-derived corrections, KoLLA references, and system outputs produced by neural or prompted GEC models. At the same time, it remains compatible with standard \texttt{m2}-style evaluation, since each correction is represented as an edit over source and target spans. The contribution of the scheme is not merely a larger label inventory, but a linguistic reinterpretation of Korean word-level edits: the surface word remains the evaluation unit, while bound morphology remains visible in the error type.

\subsection{Validation of \texttt{m2} files}
\label{validation-m2}

For \texttt{m2} validation, we compare two edit representations derived from the same NIKL data. The first representation is obtained by converting the original human morpheme-level annotations into word-level \texttt{m2} edits using Algorithm~\ref{nikl-algorithm}. The second representation is generated automatically from the reconstructed source--target sentence pairs. Agreement between these two representations indicates that the conversion preserves the correction content of the original annotations while yielding word-level edits suitable for standard GEC evaluation.

We compute true positives, false positives, false negatives, precision, recall, and F$_{0.5}$ between the converted \texttt{m2} files and the \texttt{m2} files generated from the reconstructed source--target sentence pairs. 
Precision measures how many converted edits are recovered from sentence-pair comparison, while recall measures how many sentence-pair edits are recovered by the converted annotation. 
F$_{0.5}$ gives greater weight to precision, following standard GEC evaluation practice.

\begin{table}[!ht]
\centering
\footnotesize
\begin{tabular}{r ccc ccc}
\toprule
 & TP & FP & FN & Precision & Recall & F\textsubscript{0.5} \\
\midrule
\citet{yoon-etal-2023-towards} & 48035 & 7541 & 11453 & 0.8643 & 0.8075 & 0.8523 \\
Proposed \texttt{m2} & 51118 & 3662 & 4458 & 0.9332 & 0.9198 & 0.9304 \\
\bottomrule
\end{tabular}
\caption{Validation results for two \texttt{m2} representations of the NIKL L2 corpus. The comparison evaluates agreement between edits converted from the original morpheme-level annotations and edits generated automatically from reconstructed source--target sentence pairs.}
\label{m2-file-validation}
\end{table}

Table~\ref{m2-file-validation} shows that the proposed conversion yields higher agreement with the source--target edit representation. 
It increases the number of matched edits and substantially reduces both unmatched converted edits and missed sentence-pair edits.\footnote{
The NIKL L2 corpus and the \texttt{m2} files in \citet{yoon-etal-2023-towards} use eojeol-level segmentation, where Korean spacing defines the primary token unit. 
Our proposed \texttt{m2} files separate punctuation from adjacent eojeols, following multilingual \texttt{m2} processing conventions. 
Because these segmentation choices can shift edit indices, some false positives and false negatives arise from tokenization discrepancies rather than from substantive differences in grammatical correction. 
The comparison should therefore be read as measuring the coherence between reconstructed surface forms and grammatical edits, not merely exact token-index identity.
}
The higher agreement indicates that the refined representation better preserves the correction content of the original morpheme-level annotation. 
This is especially important for Korean, where word-level edits must often encode internal changes to bound functional morphology.

\section{Multi-reference augmentation of KoLLA}
\label{multi-reference-kolla}

The refined NIKL corpus provides a large-scale basis for Korean GEC, but it remains primarily a single-reference resource. This is a structural limitation for evaluation. Learner sentences often admit more than one grammatical correction, and this is especially true in Korean, where case marking, verbal endings, word order, lexical choice, and discourse-dependent ellipsis allow multiple well-formed realizations of the same intended meaning. A single-reference design can therefore penalize system outputs that are grammatical and contextually plausible but divergent from the sole reference.

To address this limitation, we augment the Korean Learner Language Analysis corpus, KoLLA, with an additional reference correction. The original KoLLA corpus was designed for the analysis of postpositional particle errors in Korean learner writing \citep{dickinson-israel-lee-2010-building,lee-dickinson-israel-2012-developing}. Its annotation scheme is particularly relevant for Korean GEC because it treats particle errors not as isolated surface deviations, but as grammatically conditioned errors involving case, semantic role marking, and discourse function. By extending KoLLA into a multi-reference GEC resource, we use a smaller but linguistically controlled learner corpus to complement the broader NIKL-based resource.

\subsection{KoLLA corpus and annotation scope}
\label{kolla-corpus-annotation-scope}

KoLLA consists of 100 essays written by Korean L2 learners. The corpus was originally organized around four learner groups: foreign beginners, foreign intermediates, heritage beginners, and heritage intermediates. This design makes it possible to examine learner writing across both proficiency and language-background dimensions. The original annotation focused on postpositional particle errors, one of the most persistent difficulties for learners of Korean. Since Korean particles attach directly to nominal hosts and express grammatical, semantic, and discourse relations, their misuse cannot be reduced to ordinary word substitution.

The original KoLLA annotation is multi-layered. Spacing and spelling errors are handled before particle errors are classified, so that particle annotation is performed over normalized local contexts. Particle errors are then categorized as omission, substitution, addition, or ordering errors. This organization reflects the grammatical dependency between Korean word segmentation, surface form, and functional morphology. It also anticipates a central problem in Korean GEC: the correction target is a surface sentence, but many errors are internal to eojeol-level morphological structure.

For the present work, KoLLA is treated as a sentence-level GEC resource. The 100 essays are segmented into learner sentences, and each sentence is paired with reference corrections. This sentence-level design aligns KoLLA with standard GEC evaluation, where systems are typically evaluated on source-correction pairs rather than full essays. At the same time, the corpus retains its learner-oriented foundation, since the source sentences come from authentic Korean L2 writing rather than artificially generated errors.

\subsection{Additional reference correction}
\label{additional-reference-correction}

The original KoLLA corpus provides one correction layer. We augment this resource by adding a second human-generated correction for each sentence, yielding a multi-reference Korean GEC dataset \citep{song-lim-park-2026-enriching}. The additional corrections were produced by a trained linguist following detailed guidelines. The annotation aimed to produce fluent and grammatical Korean while preserving the learner's intended meaning and avoiding unnecessary rewriting.

The resulting dataset provides two reference corrections for each learner sentence. The first reference derives from the original KoLLA correction layer, while the second reference introduces an independently produced correction under the revised guidelines. This design follows the motivation of multi-reference GEC corpora such as MuCGEC \citep{zhang-etal-2022-mucgec} and MultiGEC \citep{masciolini-etal-2025-multigec}, where evaluation must account for the fact that learner errors do not always have a unique correction.

For Korean, the need for multiple references is not incidental. Functional morphology can license several grammatical alternatives depending on discourse structure. A missing postposition may be corrected by adding a case marker, by changing the syntactic configuration, or by choosing a different lexical predicate. Verbal endings may vary according to clause linkage, politeness, modality, or style. Word order may be adjusted for naturalness without changing propositional content. A multi-reference corpus therefore provides a fairer evaluation target for Korean GEC systems, because it recognizes correction as a constrained space of valid outputs rather than a single canonical string.

The augmented KoLLA annotations are represented at the word level, following current practice in neural GEC, including sequence-to-sequence models and prompted correction with large language models. This choice makes the resource directly compatible with surface-form GEC systems. At the same time, the error labels follow the Korean ERRANT-style scheme introduced in Section~\ref{korean-errant-style-annotation}, so that word-level edits remain sensitive to Korean-specific phenomena such as functional morpheme errors, word boundary errors, spelling errors, and word order errors.

\subsection{Annotation variability and correction philosophy}
\label{annotation-variability-correction-philosophy}

The two KoLLA reference layers differ not only in individual correction choices, but also in annotation philosophy. The original KoLLA corrections may use information from the broader essay context to produce a more explicit or contextually natural sentence. This is appropriate for discourse-sensitive learner corpus annotation, where the annotator may recover omitted material from the surrounding text. In contrast, the added reference follows a more minimal sentence-level correction principle: it corrects grammatical errors required for well-formedness while preserving the learner's sentence as much as possible.

This distinction is important because both principles are linguistically legitimate, but they support different evaluation settings. Context-sensitive correction is appropriate when the task is essay-level rewriting or pedagogical feedback over discourse. Minimal intervention is better aligned with sentence-level GEC evaluation, where the system is expected to correct the given sentence without introducing unnecessary content. A multi-reference design allows these perspectives to coexist, rather than forcing all valid corrections into a single editorial standard.

The contrast is especially visible in Korean because argument omission and discourse recoverability are pervasive. A sentence may be grammatical without an overt object if the object is recoverable from context. An annotator who prioritizes discourse completeness may insert the object, while an annotator who prioritizes minimality may leave the sentence unchanged. Under single-reference evaluation, one of these choices would be treated as incorrect. Under multi-reference evaluation, both can be represented as valid, provided that each satisfies the relevant annotation principle.

\subsection{Examples of divergent valid corrections}
\label{examples-divergent-valid-corrections}

Figure~\ref{korean-m2} illustrates two representative sources of divergence in the augmented KoLLA annotations. The first concerns lexical ambiguity. In Figure~\ref{korean-m2}(a), the learner sentence contains 영어 안내관 있어요 \textit{yeongeo annaegwan isseoyo}. One correction analyzes 안내관 \textit{annaegwan} as a person-denoting noun, yielding 영어 안내관이 있어요 \textit{yeongeo annaegwan-i isseoyo} (`there is an English guide'). The other correction treats the intended expression as 안내 \textit{annae} (`guidance' or `information'), yielding 영어 안내가 있어요 \textit{yeongeo annae-ga isseoyo}. Both corrections are grammatical, but they resolve the learner's intended meaning differently.

The second case concerns contextual enrichment. In Figure~\ref{korean-m2}(b), the sentence 저는 참 좋아해요 \textit{jeo-neun cham joahaeyo} (`I really like it') is grammatical as a sentence with an implicit object. One reference inserts 유미 씨를 \textit{Yumi ssi-reul} (`Yumi-\textsc{acc}') from the surrounding context, producing a more explicit sentence. The other reference leaves the sentence unchanged under minimal intervention, since no sentence-internal grammatical correction is required. This contrast shows that multi-reference annotation captures not only alternative forms, but also different assumptions about the scope of correction.

\begin{figure}[!ht]
\centering
{
\footnotesize{
\begin{tabular}{ ll  } 
(a)& \texttt{S $_{0}$그리고 $_{1}$큰 $_{2}$도시마다 $_{3}$영어 $_{4}$안내관 $_{5}$있어요 $_{6}$.}\\
&\texttt{A 4 5|||R:NOUN -> NOUN+ADP|||안내관이|||REQUIRED|||-NONE-|||0}\\
&\texttt{A 4 5|||R:SPELL|||안내가|||REQUIRED|||-NONE-|||1}\\
~&\\
(b)& \texttt{S 저는 참 좋아해요 .}\\
&\texttt{A 1 1|||M:NOUN NOUN|||유미 씨를|||REQUIRED|||-NONE-|||0}\\
&\texttt{A -1 -1|||noop|||-NONE-|||REQUIRED|||-NONE-|||1}\\
\end{tabular}
}
}
\caption{Examples of divergent annotation strategies in the Korean \texttt{m2} file. 
(a) The ambiguous learner form 안내관 \textit{annaegwan} (`guide') yields two valid corrections: one interpreting it as a person-denoting noun and adding the subject marker, and the other treating it as an erroneous form for 안내 \textit{annae} (`information' or `guidance'). 
(b) A contextually under-specified but grammatical sentence is corrected either by adding discourse-recoverable material or by applying minimal intervention.}
\label{korean-m2}
\end{figure}

These examples show why KoLLA is a useful complement to the refined NIKL corpus. NIKL supports large-scale reconstruction and annotation refinement, while KoLLA makes correction variability explicit. The augmented KoLLA corpus therefore provides a controlled testbed for evaluating whether Korean GEC systems can produce corrections that are not only close to a single reference, but acceptable within a linguistically motivated space of valid corrections.

\section{Experiments}
\label{experiments-results}

We evaluate the refined Korean GEC resources through downstream correction experiments. 
The preceding validation showed that the reconstructed NIKL targets are more fluent and that the reconstruction procedure preserves surface realization with high fidelity. 
We now examine whether these resource-level improvements translate into model-level gains. 
The experiments cover three settings: KoBART replication on refined NIKL, single- and multi-reference KoLLA evaluation, and prompted correction with generative language models.

\subsection{KoBART replication on refined NIKL}
\label{kobart-replication-refined-nikl}

We evaluate whether the refined NIKL targets improve downstream grammatical error correction under the KoBART setting of \citet{yoon-etal-2023-towards}. 
The model takes a learner sentence as input and generates the corresponding corrected sentence. 
To isolate the effect of the resource, we keep the model architecture, training protocol, and decoding setting unchanged, and replace only the target side with the reconstructed surface corrections introduced in this paper. 
We refer to the replicated model trained on the refined targets as KoBART+.

\begin{table}[!ht]
  \centering
  \footnotesize
  \begin{tabular}{lccc cccc}
  \toprule
   & TP & FP & FN & Precision & Recall & F\textsubscript{0.5} & GLEU \\
  \midrule
  KoBART & 2555 & 2418 & 5643 & 0.5138 & 0.3117 & 0.4548 & 49.82 \\
  KoBART+ & 2731 & 2186 & 5467 & 0.5554 & 0.3331 & 0.4900 & 50.51 \\
  \bottomrule
  \end{tabular}
  \caption{Performance of KoBART and KoBART+ on the refined NIKL dataset.}
  \label{model-replication}
\end{table}

Table~\ref{model-replication} shows that KoBART+ improves over the KoBART baseline under the same model setting. 
The gain is observed in both edit-based evaluation and GLEU, indicating that the refined targets improve edit selection and sentence-level similarity to the reference corrections. 
Since the model architecture and training procedure are unchanged, this improvement can be attributed to the refined supervision signal rather than to changes in the correction model. 
The result supports the role of target reconstruction as a resource-level factor in Korean GEC: incorrectly restored targets can distort both training and evaluation, whereas morphologically constrained reconstruction provides a cleaner basis for learning word-based corrections.

\subsection{Single-reference and multi-reference KoLLA evaluation}
\label{single-multi-reference-kolla-evaluation}

The KoLLA experiments evaluate how reference multiplicity affects Korean GEC scoring. 
KoBART+ is first evaluated separately against the original correction layer, denoted Reference 1, and the newly added correction layer, denoted Reference 2. 
It is then evaluated in a multi-reference setting, where an output can match either reference. 
We also compare KoBART and KoBART+ under the same multi-reference condition to test whether the effect of refined NIKL supervision carries over to KoLLA.

\begin{table}[!ht]
\centering
\footnotesize
\begin{tabular}{llccc}
\toprule
&  & \textbf{Precision} & \textbf{Recall} & \textbf{F\textsubscript{0.5}} \\
\midrule
KoBART+ & Reference 1 & 0.2948 & 0.2019 & 0.2700 \\
        & Reference 2 & 0.3293 & 0.2158 & 0.2980 \\
        & Multi-reference & 0.3937 & 0.3251 & 0.3778 \\
\midrule
KoBART  & Multi-reference & 0.3797 & 0.3197 & 0.3660 \\
\bottomrule
\end{tabular}
\caption{Evaluation of KoBART and KoBART+ on KoLLA under single-reference and multi-reference settings.}
\label{multi-reference-evaluation}
\end{table}

Table~\ref{multi-reference-evaluation} shows that single-reference evaluation is sensitive to the choice of reference layer. 
The same KoBART+ outputs receive different scores against Reference 1 and Reference 2, indicating that each reference captures different parts of the valid correction space. 
The higher score against Reference 2 does not imply that this layer is intrinsically better; rather, it shows that the added reference recovers acceptable corrections that are not represented in the original layer.

The multi-reference setting reduces this reference-dependent penalty. 
When the two references are evaluated jointly, KoBART+ obtains substantially higher precision, recall, and F$_{0.5}$ than in either single-reference condition. 
This gain should not be interpreted merely as a score increase. 
It shows that outputs rejected by one reference may still correspond to another valid correction, a common situation in Korean because of argument omission, functional morphology, word order flexibility, and lexical choice.

The comparison between KoBART and KoBART+ under the same multi-reference condition further supports the effect of refined NIKL supervision. 
KoBART+ improves over KoBART in precision, recall, and F$_{0.5}$, suggesting that the reconstructed targets provide a cleaner training signal even when evaluation is performed on a separate multi-reference resource. 
The two effects are therefore complementary: refined targets improve model learning, while multiple references provide a fairer evaluation space for acceptable Korean corrections.

\subsection{Prompted GEC evaluation with generative language models}
\label{prompted-gec-large-language-models}

We further evaluate NIKL and KoLLA as testbeds for prompted grammatical error correction with generative language models. 
For NIKL, outputs are evaluated against the refined single-reference annotations. 
For KoLLA, outputs are evaluated against the multi-reference annotations. 
This comparison is not intended as a direct dataset-to-dataset benchmark, since NIKL and KoLLA differ in composition and annotation design. 
Rather, it examines prompted correction under two evaluation settings: a refined single-reference setting and a multi-reference setting that can accept alternative valid corrections.

\begin{figure}[!ht]
\centering
\footnotesize
\begin{tabular}{p{0.92\linewidth}}
\toprule
\textbf{Prompt template used for prompted GEC evaluation} \\
\midrule
아래 한국어 문장에서 틀린 부분만 수정하고, 교정 문장 한 줄만 출력하라.\\
(`Correct only the erroneous parts in the Korean sentence below and output only one corrected sentence.')\\[2pt]
단, 문장이 이미 문법적으로 올바르면 원문을 그대로 출력하라.\\
(`If the sentence is already grammatically correct, output the original sentence unchanged.')\\[2pt]
문장: \texttt{\{sentence\}}\\
(`Sentence: \texttt{\{sentence\}}')\\[2pt]
교정문:\\
(`Corrected sentence:')\\
\bottomrule
\end{tabular}
\caption{Instruction prompt used for generative grammatical error correction.}
\label{fig:prompted-gec-prompt}
\end{figure}

\begin{table}[!ht]
\centering
\footnotesize
\begin{tabular}{lcccc}
\toprule
& \multicolumn{3}{c}{Korean model} & Multilingual model \\
\cmidrule(lr){2-4}\cmidrule(lr){5-5}
& kanana2-30b-a3b & KORMo-10b & SOLAR-10.7B-v1.0 & Qwen3-4b \\
& (instruct) & (instruct) & (instruct) & (instruct) \\
\midrule
NIKL & 0.4643 & 0.2337 & 0.1216 & 0.2824 \\
KoLLA & 0.5901 & 0.3378 & 0.1579 & 0.3638 \\
\bottomrule
\end{tabular}
\caption{Prompted GEC results of generative language models evaluated on NIKL with single-reference annotation and on KoLLA with multi-reference annotation. Results are reported as F$_{0.5}$ scores.}
\label{prompted-evaluation}
\end{table}

Using the prompt in Figure~\ref{fig:prompted-gec-prompt}, the prompted results in Table~\ref{prompted-evaluation} preserve the same model ranking across both evaluation settings: kanana2 performs best, followed by Qwen3, KORMo, and SOLAR.\footnote{The model checkpoints are \url{https://huggingface.co/kakaocorp/kanana-2-30b-a3b-instruct-2601}, \url{https://huggingface.co/Qwen/Qwen3-4B-Instruct-2507}, \url{https://huggingface.co/KORMo-Team/KORMo-10B-sft}, and \url{https://huggingface.co/upstage/SOLAR-10.7B-Instruct-v1.0}.}
All four models obtain higher F$_{0.5}$ scores on KoLLA than on NIKL. 
Because the two datasets differ, this contrast should not be interpreted solely as the effect of reference multiplicity. 
It is nevertheless consistent with the expectation that multi-reference evaluation is better suited to prompted correction, where generative models may produce grammatical outputs that diverge from a single reference in wording, lexical choice, or degree of intervention.

The comparison with supervised correction further clarifies the role of prompted generative models. 
Prompted models can produce useful corrections without task-specific training, especially when evaluation allows more than one valid reference. 
At the same time, the supervised KoBART+ model remains competitive under the KoLLA multi-reference setting. 
This suggests that prompted correction and supervised sequence-to-sequence correction benefit from the same resource-level principle: target and reference annotations should represent the space of valid Korean corrections rather than reduce grammaticality to exact agreement with a single correction path.

\section{Discussion}
\label{discussion}

The experiments show that refinement affects Korean GEC at multiple levels: target realization, edit representation, system evaluation, and reference design. 
The discussion below considers how these changes alter the interpretation of Korean GEC performance and why multi-reference correction is particularly important for Korean.

\subsection{What refinement changes in Korean GEC evaluation}
\label{what-refinement-changes}
The refined NIKL resource changes the evaluation object in Korean GEC. 
Previous evaluation depended on target sentences automatically restored from morpheme-level corrections and on edit labels derived through relatively coarse word-level mappings.
This made surface reconstruction, error classification, and system scoring difficult to separate: an error in target realization could be counted as a system error, while an underspecified edit label could obscure the grammatical nature of the correction. 
The refined pipeline reduces this instability by improving target sentence realization and by aligning word-level edits more closely with the internal morphology of Korean word forms.

The gain in target fluency shows that reconstruction is not a peripheral preprocessing step. 
Since the NIKL corpus provides corrected morpheme sequences rather than surface target sentences, target realization determines the reference string itself. 
Incorrect contraction, missing irregular conjugation, or category-insensitive rule application can introduce non-learner errors into the gold data. 
The lower perplexity of the proposed targets indicates that stricter morphologically constrained reconstruction yields more plausible Korean sentences, improving the reliability of both training data and automatic evaluation.

The \texttt{m2} validation results further show that edit conversion benefits from a linguistically explicit representation. 
Korean grammatical errors frequently involve postpositions and verbal endings, which are bound morphemes but are evaluated within word-level spans. 
Treating these forms as ordinary insertions, deletions, or replacements fails to represent the relation between a host and its functional morphology. 
By encoding functional morpheme errors as replacements over host-plus-function configurations, the proposed scheme preserves word-level compatibility while making the relevant grammatical contrast visible. 
The improved precision, recall, and F$_{0.5}$ in \texttt{m2} validation therefore reflect not only better string alignment, but also a more adequate mapping between Korean morphology and GEC edits.

This refinement also changes the interpretation of system performance. 
A model evaluated against noisy targets or coarse edit labels may appear to make errors that are artifacts of reconstruction or annotation. 
Conversely, a system may receive credit for matching a defective reference. 
By improving both target realization and edit classification, the proposed evaluation reduces these distortions. 
The KoBART+ results suggest that resource refinement can yield measurable gains even without changing the model architecture. 
In this respect, the contribution is not simply a new annotation inventory, but a more stable evaluation foundation for Korean GEC.

\subsection{Why multi-reference correction matters for Korean}
\label{why-multi-reference-matters}

Multi-reference correction is particularly important for Korean because grammatical well-formedness is often compatible with several surface realizations. 
Korean allows extensive argument omission, flexible constituent order, and multiple choices of case marking and verbal endings depending on discourse structure, information status, politeness, and stylistic preference. 
A learner sentence may therefore have more than one valid correction, even when the intended meaning is held constant. 
Single-reference evaluation collapses this space of acceptable corrections into one string, turning legitimate variation into apparent system error.

The augmented KoLLA evaluation makes this problem visible. 
The difference between the two single-reference settings shows that evaluation scores are sensitive to the choice of reference. 
The higher scores under multi-reference evaluation indicate that system outputs often match one valid correction even when they diverge from another. 
This is not merely a matter of giving systems more chances to match. 
It reflects the linguistic fact that Korean correction is frequently underdetermined by the source sentence alone. 
Lexical ambiguity, recoverable arguments, and alternative functional morphology can all produce distinct but acceptable corrections.

The contrast between context-sensitive correction and minimal intervention is central here. 
In discourse-level learner corpus annotation, adding recoverable material from surrounding context may produce a more explicit and pedagogically useful correction. 
In sentence-level GEC evaluation, the same addition may be unnecessary if the sentence is already grammatical. 
Both correction principles are valid, but they instantiate different task assumptions. 
A multi-reference resource allows these assumptions to be represented rather than forcing them into a single annotation policy.

This point is especially relevant for prompted correction with large language models. 
Such models often produce fluent outputs that are not literal matches to a single reference. 
In Korean, these differences may involve acceptable paraphrase, altered information structure, or alternative but grammatical functional morphology. 
The higher prompted GEC scores on KoLLA are consistent with the view that single-reference scoring can underestimate the acceptability of many model outputs, although NIKL and KoLLA also differ in composition and annotation design. 
Multi-reference evaluation therefore provides a more realistic testbed for current GEC systems, particularly when the system is capable of producing fluent corrections beyond narrow edit reproduction.

\section{Conclusion and future perspectives}
\label{conclusion}

This paper refined Korean grammatical error correction resources by addressing three representational bottlenecks: surface target realization, Korean-specific edit annotation, and single-reference evaluation. 
Starting from the NIKL L2 learner corpus, we reconstructed more reliable surface target sentences from morpheme-level corrections and converted the original annotations into word-level \texttt{m2} edits that remain sensitive to Korean bound morphology. 
The validation results show improved target fluency, high-fidelity surface reconstruction, and stronger agreement between converted and automatically generated edit representations.

We also introduced a Korean ERRANT-style annotation scheme that preserves the MRU core while extending it to Korean-specific phenomena. 
Functional morpheme errors are treated as replacements over host-plus-function configurations, spelling errors are represented conservatively as \texttt{R:SPELL}, and word boundary and word order errors are defined through explicit string- and set-based conditions. 
This scheme keeps Korean GEC compatible with standard edit-based evaluation while avoiding the treatment of Korean word forms as morphologically opaque strings.

The augmentation of KoLLA with an additional reference correction addresses a central limitation of single-reference GEC evaluation. 
Korean learner sentences often allow multiple valid corrections because of argument omission, functional morphology, word order flexibility, and discourse-dependent interpretation. 
Multi-reference evaluation captures part of this variability and provides a fairer basis for assessing both supervised and prompted GEC systems. 
The experiments show that refined targets improve KoBART-based correction under the same model setting, and that multi-reference KoLLA evaluation reduces the risk of penalizing linguistically valid generations that differ from a single gold correction.

Several directions follow from these results. 
First, target reconstruction should be strengthened through broader lexical validation and more systematic treatment of irregular morphology. 
Although the proposed rules reduce erroneous realization, Korean inflection includes lexicalized and category-sensitive patterns that require richer lexical information. 
Second, Korean GEC evaluation should move toward tighter integration between word-level scoring and morpheme-level analysis. 
The present scheme remains word-based for compatibility with standard GEC evaluation, but a future morpheme-aware framework could represent bound functional morphology more directly.

Third, spelling and reference variation require finer modeling. 
The present scheme treats spelling errors conservatively, but Korean learner spelling may reflect phonological confusion, morphophonological misanalysis, syllable-level substitution, or orthographic convention errors. 
Likewise, the two-reference KoLLA setting improves coverage but does not exhaust the space of valid corrections. 
Future multi-reference annotation should make correction principles explicit, distinguishing, for example, minimal sentence-level intervention from discourse-sensitive enrichment.

Finally, Korean GEC should be extended beyond L2 learner data and evaluated in relation to pedagogical usefulness. 
Native Korean writing raises different correction issues, including spacing, spelling, register, punctuation, and norm-sensitive variation. 
At the same time, learner-oriented GEC should be judged not only by edit overlap, but also by whether corrections preserve learner intent, avoid unnecessary normalization, and support interpretable feedback.

The broader implication is that Korean GEC cannot be made robust by model development alone. 
Evaluation depends on the linguistic adequacy of the reference data, the correctness of surface realization, and the granularity of edit labels. 
By refining these foundations, this work provides a more reliable basis for Korean learner corpus annotation, system training, and cross-linguistic comparison in grammatical error correction.

\paragraph{Acknowledgments}
We gratefully acknowledge the National Institute of Korean Language for providing access to the NIKL corpus. 
We also thank Markus Dickinson, Ross Israel, and Sun-Hee Lee for creating and sharing the KoLLA corpus, which provides an important foundation for this work.

\paragraph{Funding statement}
This work was supported by the Institute of Information \& Communications Technology Planning \& Evaluation (IITP) grant funded by the Korea government (MSIT) (No. RS-2025-25441313, Professional AI Talent Development Program for Multimodal AI Agents).


\end{document}